\documentclass[review,11pt]{elsarticle}

\journal{Fundamenta Informaticae}

\usepackage{amsmath}
\usepackage{amssymb}
\usepackage{multirow}
\usepackage{amssymb}
\usepackage{graphicx}
\usepackage{array}
\usepackage{subfigure}
\usepackage{epsfig}
\usepackage{epstopdf}

\DeclareGraphicsExtensions{.eps}

\newtheorem{definition}{Definition}

\newtheorem{theorem}{Theorem}

\date{}

\begin{document}

\title{An Affine moment invariant for multi-component shapes}

\author{Jovi\v{s}a \v{Z}uni\'{c}\footnote{Mathematical Institute, Serbian Academy of Sciences, Belgrade, Serbia \newline \hspace*{0.5cm}  e-mail:    \ jovisa\_zunic@mi.sanu.ac.rs}, 
\ \ 
Milo\v{s} Stojmenovi\'{c}\footnote{Department of Informatics and Computing, 
Singidunum University, Belgrade, Serbia  \newline 
\hspace*{0.5cm}  email:   \  mstojmenovic@singidunum.ac.rs}}

\begin{abstract} We introduce an image based algorithmic tool for analyzing multi-component shapes here.
Due to the generic concept of multi-component shapes, our method can be applied to the analysis of a wide
spectrum of applications where real objects are analyzed based on their shapes - i.e. 
on their corresponded black and white images. The method allocates a number to a shape, herein called a multi-component shapes measure. This number/measure is invariant with respect to affine transformations and is established based on the theoretical frame developed in this paper.   
In addition, the method is easy to implement and is robust (e.g. with respect to noise). 

We provide two small but illustrative examples related to aerial image analysis and galaxy image analysis. Also, we provide some synthetic examples for a better understanding of the measure 
behavior. \\[0.0cm]
\end{abstract} 

\endfrontmatter

 \section{Introduction}\label{intro}
 
A shape is an object characteristic that can be described by several numerical characteristics.
These numerical characteristics, herein called {\it shape measures}, are employed
in many computer vision, image processing, and pattern recognition tasks 
\cite{knjiga-1,knjiga-2}. Shape measures are often designed to evaluate certain 
shape properties, like {\it elongation} \cite{aktas,anastazia}, {\it convexity}
\cite{convexity-1,convexity-2},  {\it tortuosity} \cite{tortuosity}, and many more.
Some of these measures are generic, and aim to satisfy some specific properties 
such as rotational \cite{hu,rotacija} or affine \cite{affine,suk} 
invariance, for example. It is worth mentioning that a range of popular shape descriptors 
exist, and are extensively applied: algebraic \cite{hu}, geometric 
\cite{xu}, logical \cite{logical}, 
fractal ones \cite{fractal}, 
and so on. In most cases the shape measures are combined and used
together to achieve a better performance \cite{expert-1}.

In this paper we define a measure applicable to multi-component shapes, which is
invariant with respect to Affine transformations. We examine
multi-component shapes in detail since they are not yet intensively studied in 
literature, whereas there exists a lot of the work related to Affine invariants, 
both in theory and practice.

\subsection{Multi-component shapes}
\begin{figure}[t!]
\begin{center}
\begin{tabular}{cccc}
             \includegraphics[width=3.0cm]{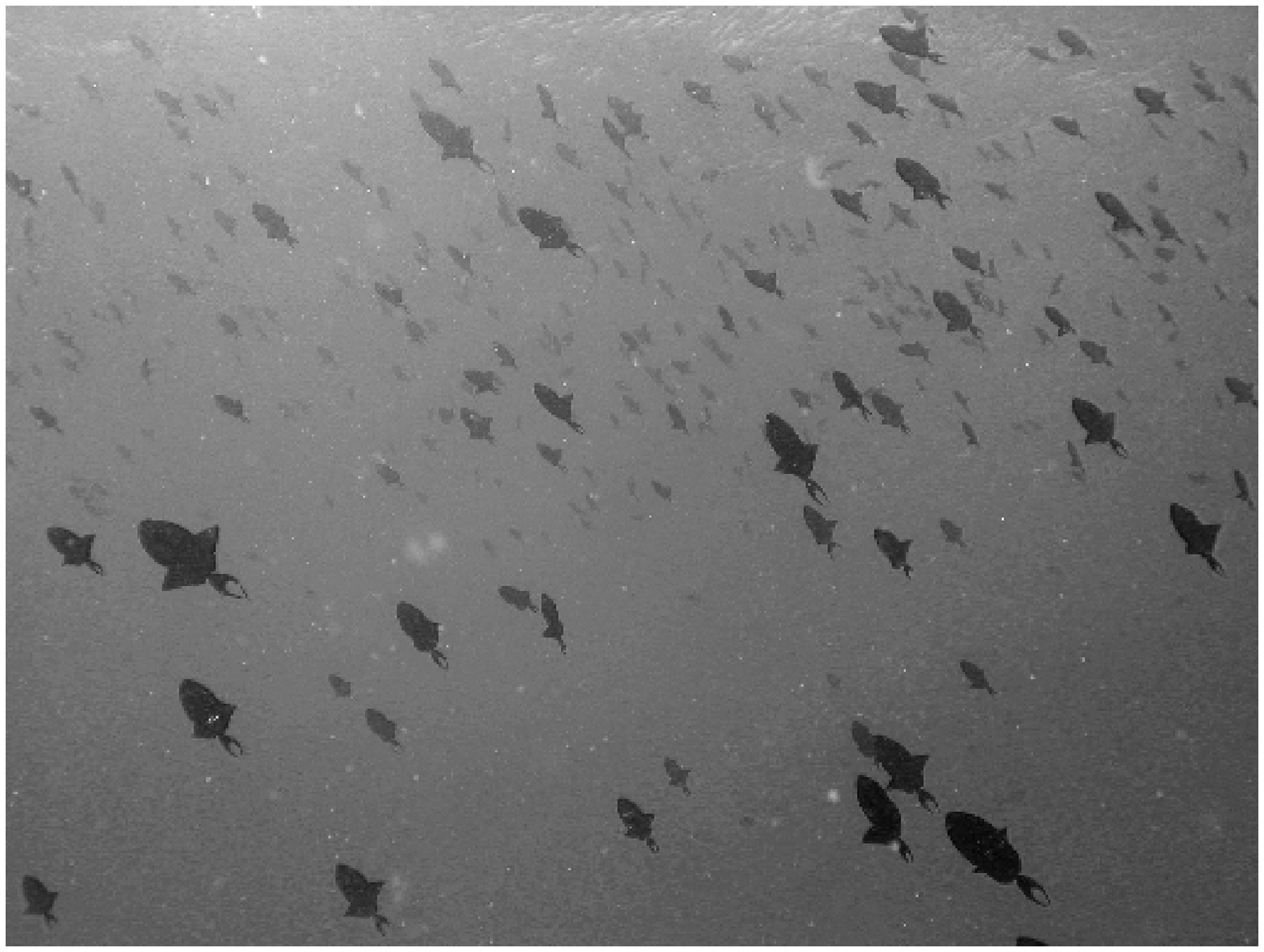}  &  
						\includegraphics[width=2.6cm]{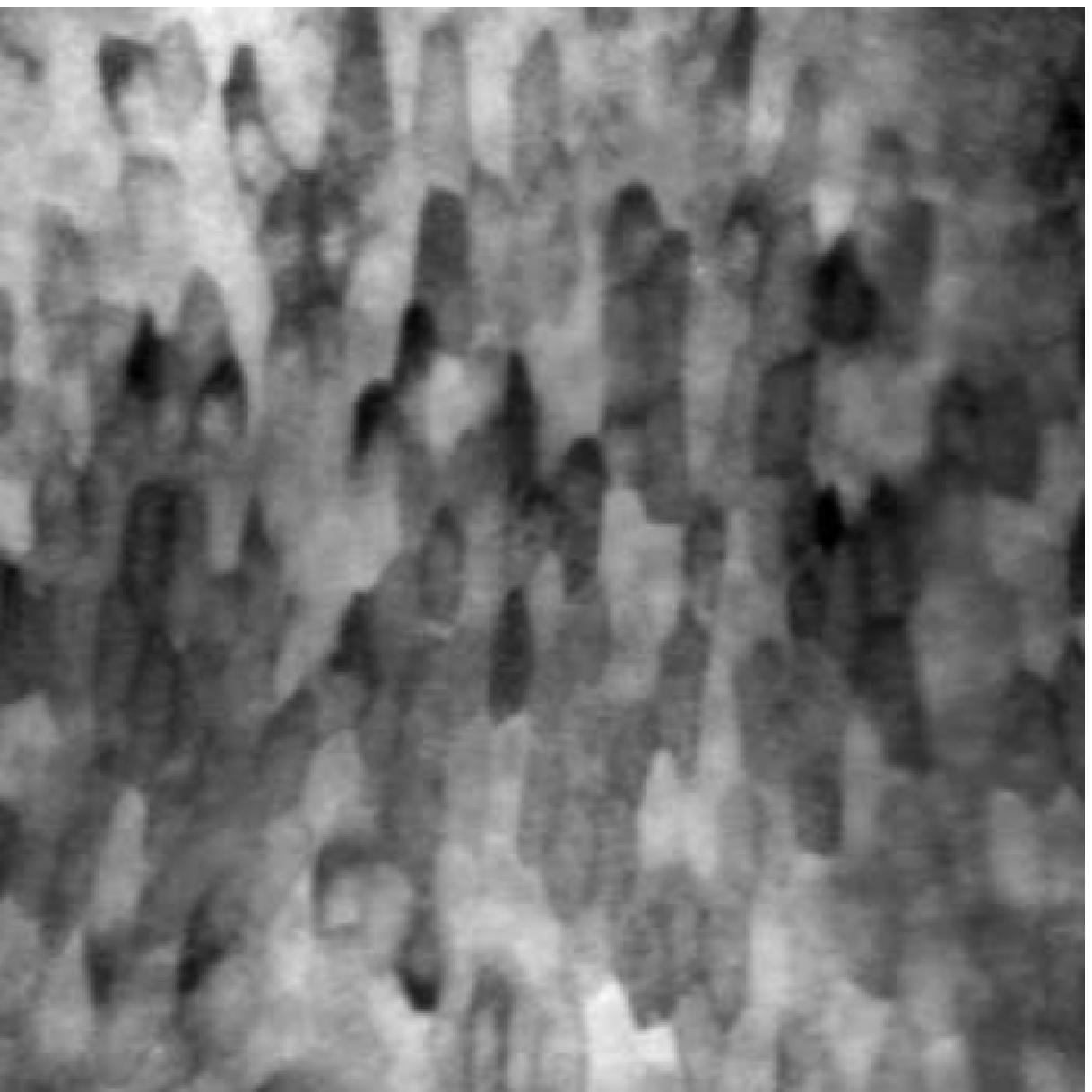}   &  
             \includegraphics[width=2.2cm]{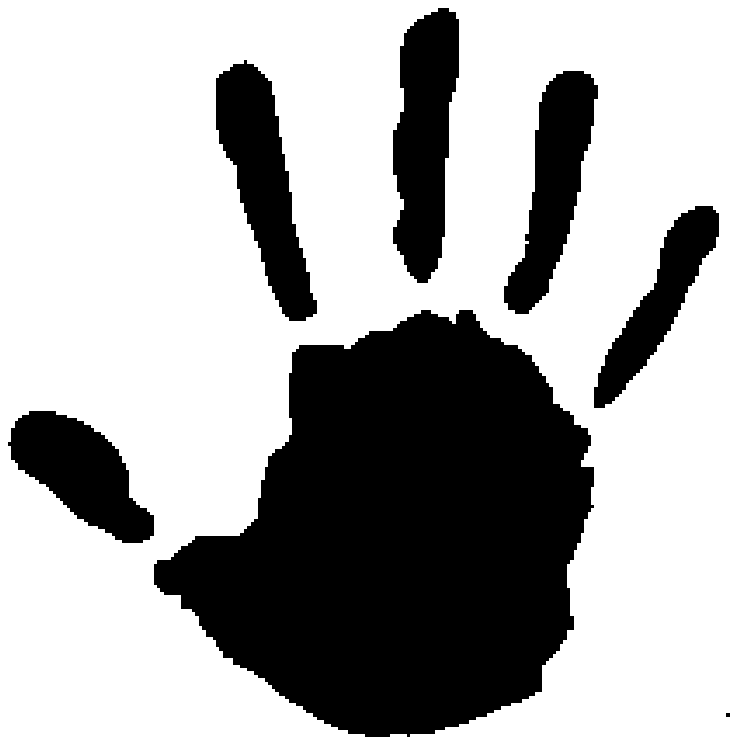}  & 
						\includegraphics[width=2.3cm]{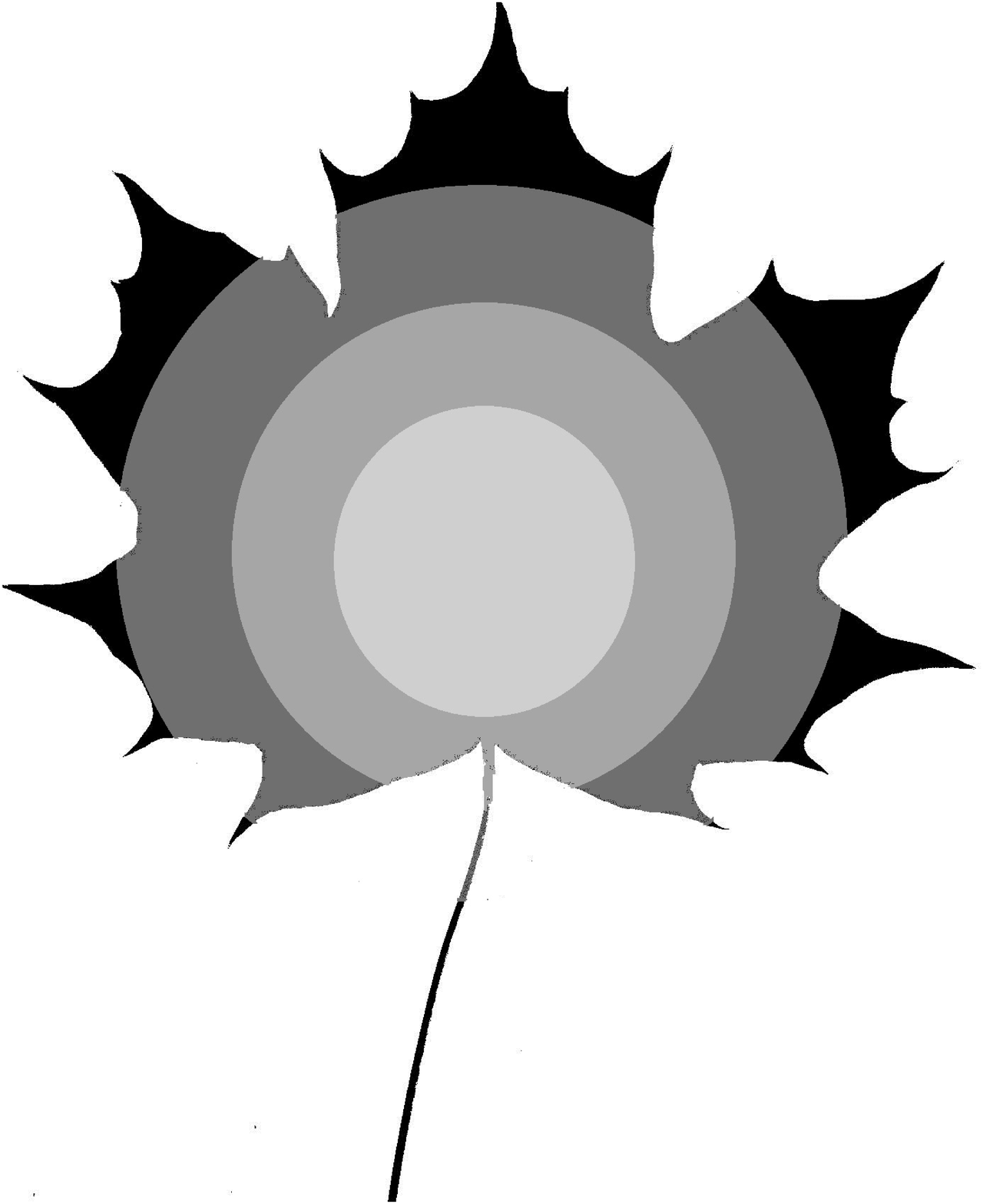} \\
             (a) &   (b)  &   (c)   &    (d)  
\end{tabular}       \\[0.9cm]
\begin{tabular}{cc}
  \includegraphics[width=1.5cm]{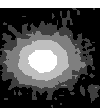}    &  
  \includegraphics[width=9.5cm]{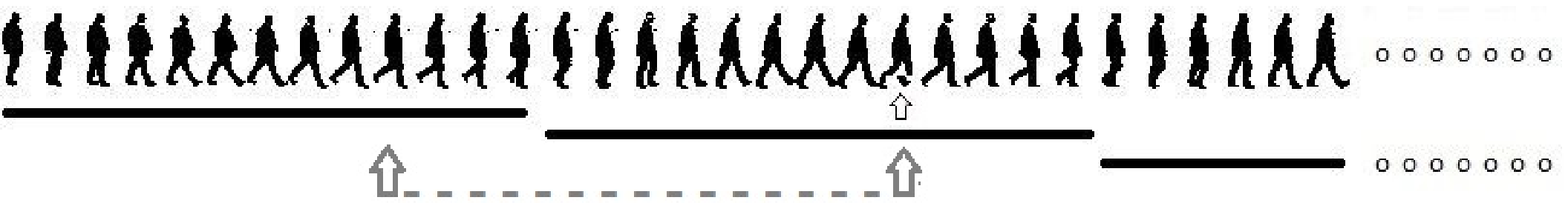} \\
     (e)   \ \  & \ \     (f)
\end{tabular}
\end{center}
\caption{(a) School of Fish (b)   Embryonic
tissue with indistinct cell boundaries  (c) Palm-print
(d) leaf shape decomposed by concentric circles (e) galaxy shape obtained with a
thresholding method \cite{otsu}
 (f) Human gait --  considered as a 13-component shape,
whose components are the appearances of a walking person in a
sequence of 13 consecutive frames.} 
\label{illustrations}
\end{figure}

The concept of multi-component shapes has been presented in \cite{zr-ijcv}. This is a very generic 
concept that allows us to segment a single object onto components, to group objects into a
multi-component shape to suit a particular application. Some examples are given in 
Fig.\ref{illustrations} and in the Experimental illustration section. Notice that some shape 
segmentation results can be natural (Fig.\ref{illustrations}(c)), while some can be more
artificial (e.g. Fig.\ref{illustrations}(d,e)); some components can be connected, some
not ((e.g. Fig.\ref{illustrations}(d,e))). Basically, there are no real restrictions on 
categorizing a shape as a being multi-component. Of course, a common requirement is that 
the selected segmentation well suits the desired application. 

\begin{figure}[htb]
   \subfigure[]{\includegraphics[width=3.6cm]{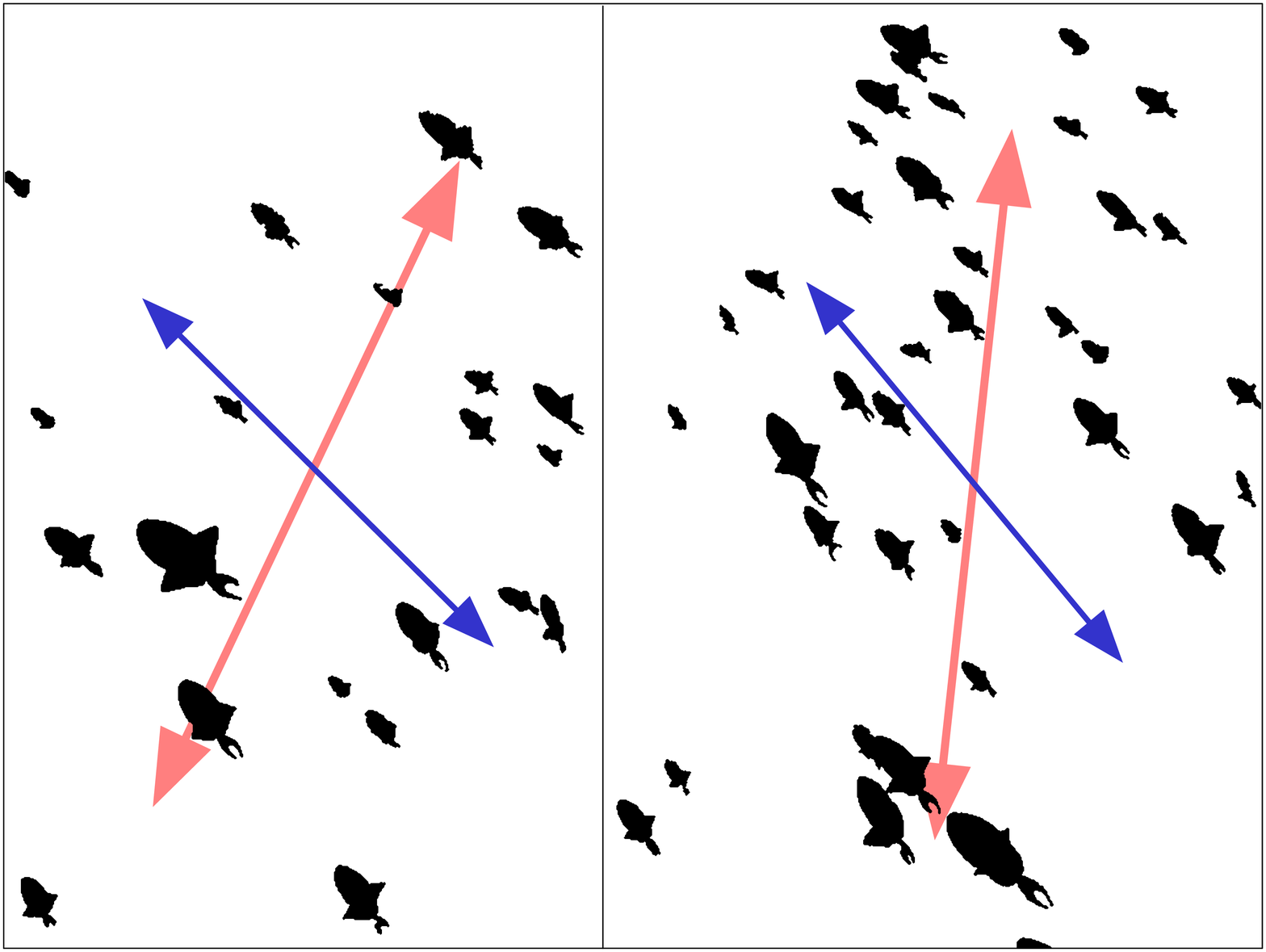} }
   \hfill
   \subfigure[]{\includegraphics[width=3.6cm]{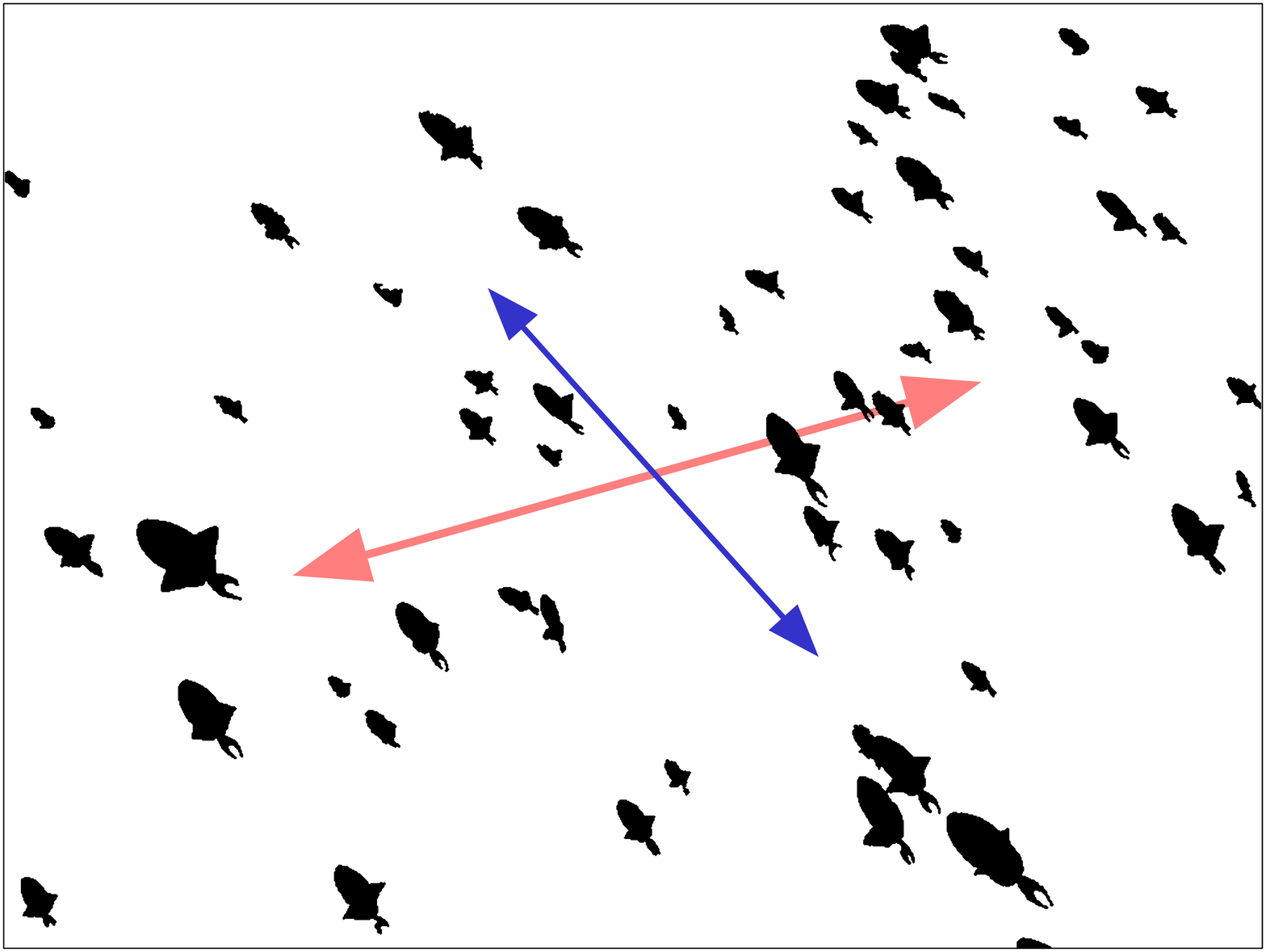}}
	\hfill
   \subfigure[]{\includegraphics[width=3.6cm]{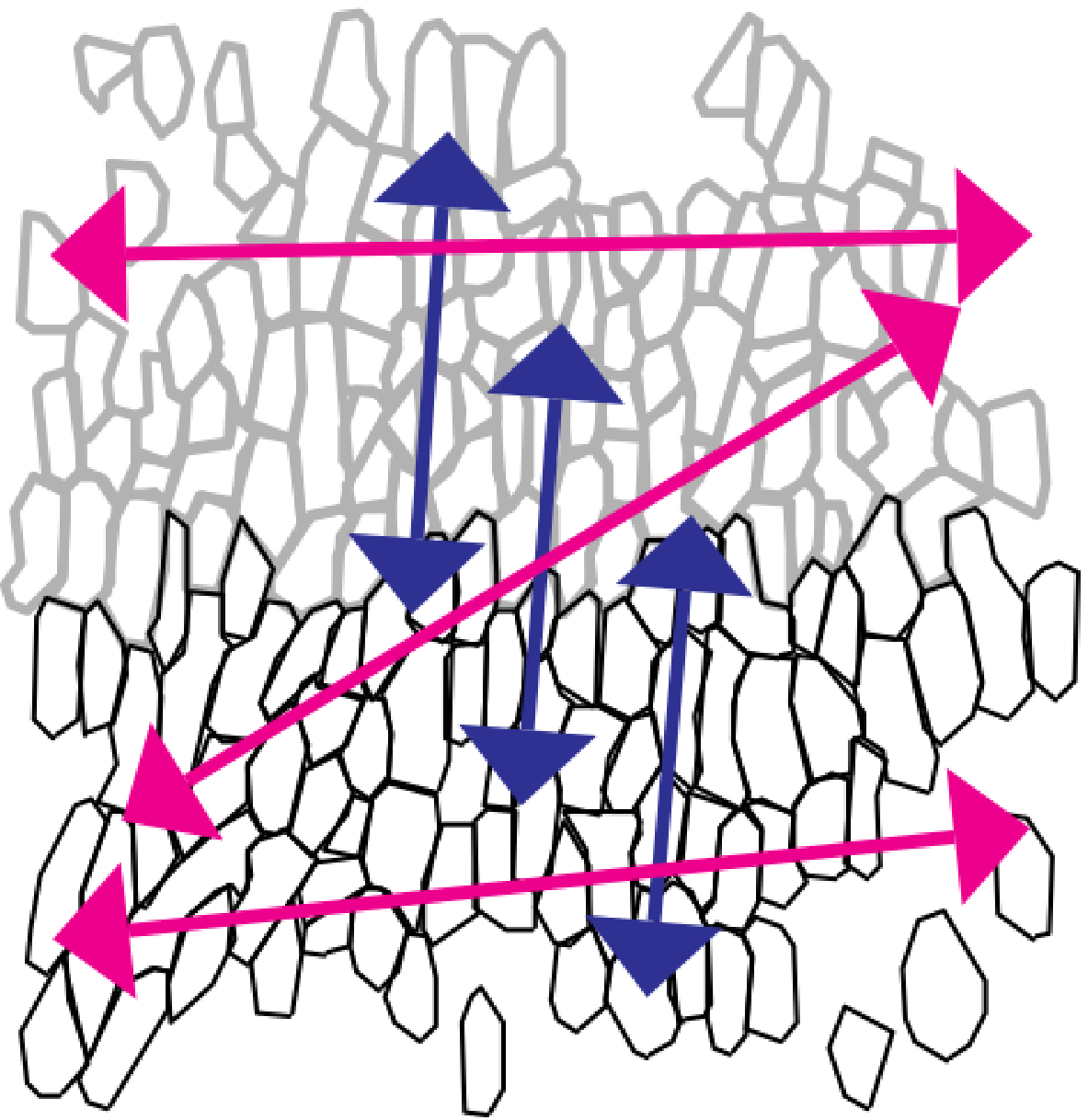}}
\caption{Orientations computed for (a) the separate left and right halves of the image,
and also for (b) the complete image.
The shorter dark blue arrows correspond to the 
shape orientations computed for the three sets of data according to our multi-component approach.
In comparison, the long light red arrows correspond to the traditional 
method for computing orientation, in which each of the three sets of data are considered as 
representing single-component shapes (each containing multiple fish).
}
\label{polovine}
\end{figure}

Examples (a) and (b), in Fig.\ref{illustrations} 
are real images of a school of fish and an embryonic tissue, each of which have
recognizable components. Their extracted components are in Fig.\ref{polovine}.
Shape (c) in Fig.\ref{illustrations} represents a natural "decomposition" of a palm print
while shapes (d) and (e), in the same figure represent an artificial or inaccurate decomposition
of a leaf shape \cite{rhouma} and a galaxy shape, respectively, obtained using a thresholding method.

As expected, multi-component shapes are expected to have some properties that are not typical 
of single shapes. An example of such a property is the orientation of multi-component shapes. Indeed, in the case of   
relatively large numbers of components, it is expected that the orientation of a multi-component
shape does not depend much on a "window" used for the orientation computation 
\cite{zr-ijcv,anisotropy}. Two examples are in found in Fig.\ref{polovine}. The computed orientations 
of both halves, as well as the shape as a whole almost coincide for both (a) and (c) in
Fig.\ref{polovine}.
 
There already exist shape descriptors whose measures 
are only applicable for multiple component shapes. Some examples are anisotropy and 
dis-connectedness of multi-component shapes \cite{anisotropy,zri}.

\section{Preliminaries}\label{Prelim}

Here we define the basic terms and introduce the notation used in this paper. 
We also give some well-known facts used for our derivations. \\[0.3cm]
$\bullet$ The shape is a basic object property (like color or texture, for example). As such, shape does not need a formal definition. Shape is represented by a planar region, 
usually displayed as a binary (black-white) image. 

$\bullet$ A multi-component shape $S = S_1 \cup S_2\cup \ldots \cup S_n$, having components 
$S_1, \;  S_2, \; \ldots, \; \makebox{and}, \; S_n$,
is represented by $n$ regions, corresponding to its components.  

Note that the regions  representing a component $S_i$ do not need to be connected, in a 
topological sense. \footnote{Formally, an $n$-component shape 
$S=S_1\cup S_2 \cup \ldots \cup S_{n}$,  can be  defined formally
as a mapping $F((x,y))$: \ \ $S\ \rightarrow \ \{1,2,\ldots,n\}$. 
All of the points $(x,y)$ with the same assigned value $F((x,y))$,  
belong to the same component, i.e. the $k$-th component $S_k$, where
$1\leq k \leq n,$ is defined as  \ \ \
$     
S_k  \ = \ \{ (x,y) \ | \ F((x,y))=k\} \ = \  F^{-1}(k).
$}
There is no formal restriction on  how a given shape can be 
decomposed and presented as a multi-component shape. From a practical view-point such a
decomposition should be meaningful in order to be beneficial for the desired application.
  
$\bullet$ Two shapes are considered equal if their set difference has an area equal to zero.

$\bullet$ The {\it geometric moment}, or simply {\it  moment}, 
$m_{p,q}(S)$ of a given planar shape $S$ is defined as
\begin{equation}
m_{p,q}(S)  = \int\limits_S\!\!\!\!\int x^p y^q  dx\,   dy.
\label{moments-def}
\end{equation}
The moment $m_{p,q}(S)$  has the {\it order}  equal to $p+q$. 
Obviously,  $m_{0,0}(S),$ equals the area of $S$, while 
the moments $m_{1,0}(S)$ and $m_{0,1}(S)$ are used to define the {\it shape centroid},
denoted as  $\ \left(x_c(S), \ y_c(S)\right) \ $ and formally defined as 
\begin{equation} 
\left(x_c(S),\ y_{c}(S)\right)\; = \; 
\left(\frac{m_{1,0}(S)}{m_{0,0}(S)},\  \frac{m_{0,1}(S)}{m_{0,0}(S)}\right).
\label{centroid-def}
\end{equation}

$\bullet$ The normalized moments $\mu_{p,q}(S)$ are defined as
\begin{equation}
\mu_{p, q}(S) \ = \ \frac{1}{m_{0,0}(S)^{(p+q+2)/2}} \cdot 
\int\limits_S\!\!\!\int
\left(x - x_c(S)\right)^{p}
\left(y - y_c(S)\right)^{q} dx   dy   
\label{normalized-moments-def}
\end{equation}

$\bullet$ Normalized moments $\mu_{p,q}(S)$ are translation and scaling 
invariant by definition. 
Both invariances are required in shape based tasks because shape properties 
do change under scaling and translation transformations. 

$\bullet$ Many more moment invariants are used in shape based image analysis tasks.
 In his seminal work \citep{hu}, Hu introduced seven 
quantities which are rotational invariants. Hu used algebraic reasoning, but later on 
Xu and Li \citep{xu} showed that Hu invariants are actually geometric invariants, and can be 
derived by considering certain geometric primitives defined by the shape points. 
Geometric reasoning will be applied here as well.

$\bullet$ Affine moment invariants are quantities computed from a set of moments that
do not change under Affine transformation \cite{suk, suk-2}. The following Affine invariant will
be exploited here  
\begin{equation}
{\cal A}(S)   =   \mu_{2, 0}(S) \cdot  \mu_{0, 2}(S) - \mu_{1, 1}(S)^2.
\label{A-def}
\end{equation}

$\bullet$ The Affine invariant ${\cal A}(S)$ has a simple geometric interpretation. Indeed,
  the squared area of the triangle whose vertices are points 
$(0,0),$ $(x,y)$, and $(u,v)$ is 
\begin{equation}
\frac{1}{4}\cdot (x\cdot v - y \cdot u)^2.
\label{triangle-area}
\end{equation}
So, for a given shape $S$, having an area equal to $1$, and all the triangles $\Delta ABC$ such
that $A=(x,y)\in  S$ and $B(u,v)\in S$  
we easily obtain that a half of ${\cal A}(S)$ equals the total integral of the squared  areas of triangles $\Delta ABC$:
\begin{eqnarray}
&&
\iint\limits_{(x,y)\in S}  \ \ 
\iint\limits_{(u,v)\in S}  \ \ 
\frac{1}{4}   \cdot   (x\cdot v - y \cdot u)^2 dx\ dy\ du\ dv \nonumber \\[0.3cm]
&=& \frac{1}{2} \cdot \left(\mu_{2, 0}(S) \cdot  \mu_{0, 2}(S) - \mu_{1, 1}(S)^2\right) 
= \frac{1}{2} \cdot {\cal A}(S).
\label{equality-a}
\end{eqnarray} 
 
Such a simple geometric interpretation \cite{jdjmiv,xu} has been used in \cite{jdjmiv} to 
establish a shape interpretation of ${\cal P}(S)$. This interpretation of ${\cal P}(S)$ does not
allow a simple extension to multi-component shapes. In the next section 
we will give a modified interpretation of ${\cal P}(S)$, which is extendable to multi-component shapes.


\section{A new Affine invariant for multi-component shapes} 

As mentioned, we aim to develop a multi-component shape
measure that is invariant with  respect to Affine transformations. Our first idea was 
to exploit an existing Affine moment invariant ${\cal A}(S)$, as given  in 
(\ref{equality-a}), defined for single-component shapes \cite{suk}.
Such a measure has a nice geometric interpretation \cite{xu,jdjmiv} but is not easily extendable to 
measure multi-component shapes.\footnote{To the best of our knowledge, this interpretation is
novel, never before mentioned in literature. Since the idea is natural, it is still possible that 
it has been used and observed by others already.}
So, we had to modify the equality in (\ref{equality-a}) and derive another geometric
interpretation of the measure ${\cal A}(S).$ We show that ${\cal A}(S)$ 
(as given in (\ref{A-def})) equals the integral of the squared triangle areas, i.e. the integral
of $Area\_(\Delta ABC)^2$, where $A,$ $B,$ and $C$ vary through a given shape $S$ whose 
area is $1.$ We give the following theorem.
\begin{theorem}
Let $A=(x,y),$ $B=(u,v),$ and $C=(z,\omega)$, be points belonging to a given shape $S$, having an area equal to $1,$. 
The integral of all squared areas of triangles $\Delta ABC$ equals a quarter of ${\cal A}(S),$ or formally
\begin{eqnarray}
&&\frac{1}{6}\iint\limits_{(x,y)\in S} \ \ \iint\limits_{(u,v)\in S} \ \
 \iint\limits_{(z,\omega)\in S}
(Area\_of\_\Delta ABC)^2 dx dy du dv dz d\omega \nonumber \\[0.3cm]
&&\frac{1}{6} \ \cdot \ \iint\limits_{(x,y)\in S} \ \ \iint\limits_{(u,v)\in S} \ \ \iint\limits_{(z,\omega)\in S}
\frac{1}{4}\cdot 
\left((x-z)(v-\omega) - (y-\omega)(u-z)\right)^2 dx dy du dv dz d\omega \nonumber \\[0.3cm]
&=&\frac{1}{4}\cdot \left(\mu_{2,0}(S)\cdot \mu_{0,2}(S) - \mu_{1,1}(S)^2\right) \ = \ 
\frac{1}{4}\cdot {\cal A}(S).
\label{total-area}
\end{eqnarray}
\end{theorem}
{\bf Proof.} We do not give a complete proof here. Even though the derivation might be seen as 
a long and messy one, it is still trivial.
Here are some hints to support the derivations.
\begin{itemize}
\item The formula in (\ref{triangle-area}) for the points
$(x-z,y-\omega),$ $(u-z,v-\omega),$ and $(z-z,\omega-\omega)=(0,0)$, should be used.
\item The identity
$\iint\limits_{(x,y)\in S} \ \ \iint\limits_{(u,v)\in S} \ \ \iint\limits_{(z,\omega)\in S}
x^a y^b u^c v^d z^e \omega^f dx dy du dv dz d\omega 
= \mu_{a,b}(S) \cdot \mu_{c,d}(S) \cdot \mu_{e,f}(S)$ 
should be applied. 
\item The factor $\frac{1}{6}$ comes from the fact that three points 
$(x,y),$ $(u, v),$ and $(z, \omega),$  determine the same triangle, independently
on order they have been selected. \hfill   $\blacksquare$
\end{itemize} 
Now, we have the idea and necessary theoretical framework regarding how to define the 
affine moment invariant for multi-component shapes. Briefly, we observe all points that 
do not belong to a single shape component and the integral of the squared areas of the 
triangles determined by these points.
Rather than manipulate with all three points (i.e. observed triangle vertices) that do not
belong to the same component, we observe the squared triangle areas determined by all of the 
triplets of points that do belong to the n-component shape 
$S=S_1\cup S_2\cup \ldots \cup S_n$ and take the areas of all the triangles whose vertices 
belong to the same component. This is just for the sake of a simpler computation.

Since the shape $S$ and all of the components of $S$ should play a role in the computation
of its new Affine invariant measure, the size of all of these appearing shapes should be 
taken into account. We have the following identity (see (\ref{normalized-moments-def})), 
for the shapes whose area is not equal to $1$:
\begin{eqnarray}
& & m_{2,0}(S)\cdot m_{0,2}(S) - m_{1,1}(S)^2  \ =  \nonumber \\[0.2cm]
& & \mu_{0,0}(S)^4\cdot \left(\mu_{2,0}(S)\cdot\mu_{0,2}(S) - \mu_{1,1}(S)^2\right) 
\ = \ \mu_{0,0}(S)^4\cdot {\cal A}(S).
\label{measure-shape}
\end{eqnarray}
As mentioned, we observe the integral of squared areas of all triangles whose vertices do
belong to the same component of the n-component shape $S$
 from the integral of squared areas of all triangles
whose vertices belong to the shapes $S$. 
This quantity can be computed as
\begin{equation}
\left(m_{2,0}(S)\cdot m_{0,2}(S) - m_{1,1}(S)^2\right)  \  -  \ 
\sum_{i=1}^{i=n} \left(m_{2,0}(S_i)\cdot m_{0,2}(S_i) - m_{1,1}(S_i)^2\right)
\label{quantity-1}
\end{equation}
or equivalently  in the following form
\begin{eqnarray}
&& m_{0,0}(S_1\cup \ldots \cup S_n)^4 \cdot
\left(\mu_{2,0}(S)\cdot \mu_{0,2}(S) - \mu_{1,1}(S)^2\right)  \nonumber \\[0.2cm]
&& \qquad - 
\sum_{i=1}^{i=n} m_{0,0}(S_i)^4 \cdot 
\left(\mu_{2,0}(S_i)\cdot \mu_{0,2}(S_i) - \mu_{1,1}(S_i)^2\right). 
\label{quantity-2}
\end{eqnarray}
If we normalize the above form, i.e. divide both summands with 
$\left(m_{0,0}(S_1\cup \ldots \cup S_n)\right)^4$ we get the quantity    
\begin{equation} 
{\cal A}(S_1\cup \ldots \cup S_n) - 
\frac{1}{m_{0,0}(S_1\cup \ldots \cup S_n)^4  } \cdot
\sum_{i=1}^{i=n} m_{0,0}(S_i)^4\cdot {\cal A}(S_i). 
\label{quantity-3}
\end{equation} 
that will be used to define a new Affine measure ${\cal M}(S)$ for 
multi-component shapes.
We give the following definition.
\begin{definition}
Let an $n$-component shape $S$ be given. We define the multi-component shape measure ${\cal M}(S)$,
for such a shape $S$, by the following equality
\begin{eqnarray} 
&& {\cal M}(S)  =  {\cal M}(S_1\cup \ldots \cup S_n) \nonumber \\[0.2cm]
& = &   {\cal A}(S_1\cup \ldots \cup S_n) - 
\frac{1}{m_{0,0}(S_1\cup \ldots \cup S_n)^4  } \cdot
\sum_{i=1}^{i=n} m_{0,0}(S_i)^4\cdot {\cal A}(S_i). 
\label{quantity-4}
\end{eqnarray} 
\label{definition-1}
\end{definition}

\section{Experimental Illustrations}\label{Experiments}
\begin{figure}[h!]
\begin{center}
\begin{tabular}{c | cc}
             \includegraphics[width=3.6cm]{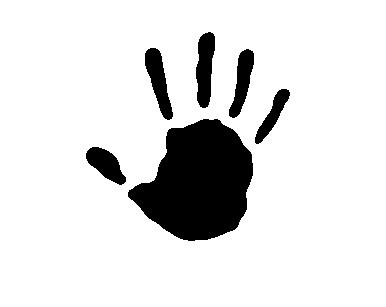}&  
					   \includegraphics[width=3.6cm]{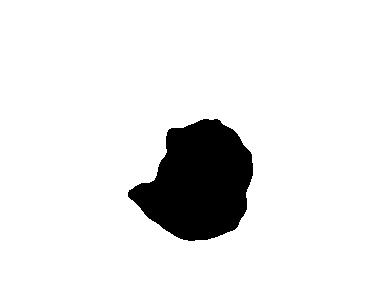}&  
             \includegraphics[width=3.6cm]{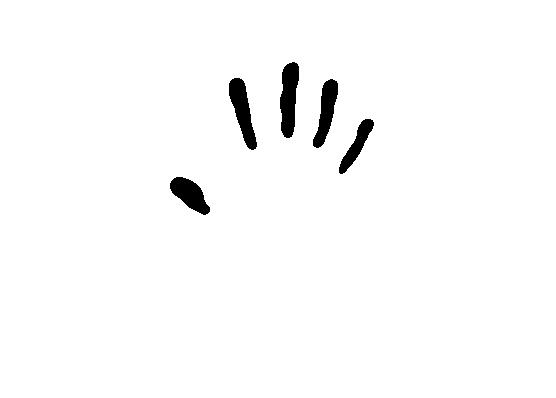}  
						\\ 
               {\bf (a):}\ {\footnotesize  A whole  hand shape} &  
							 {\bf (b):}\ {\footnotesize   The palm} &  
							 {\bf (c):}\ {\footnotesize   The fingers}   \\
							${\cal M}_{6-comp.}(S)=0.0162$ & ----- & ----- \\
							${\cal M}_{2-comp.}(S)=0.0177$ & ----- & -----\\
							----- & ${\cal M}_{1-comp.}(S)=0 $ & ${\cal M}_{5-comp.}(S)=0.1972$
\end{tabular}
\end{center}
\caption{The shape on the left is treated as being 2-component (five fingers as one components and 
the palm as the second component) and its computed measure is 
${\cal M}_{2-comp.}(S)=0.0177$. Treating the same shape as being 6-component  
(five  fingers, each a separate  component, and the 
palm, as the final component), the computed measure is ${\cal M}_{6-comp.}(S)=0.0162$. 
The measure values for 
the 1-component shape and the 5-component shape in Fig.\ref{saka}(b)\&(c) 
are \  $0$ \ and  \  $0.1972,$ \ respectively.}
\label{saka}
\end{figure}

Several experimental illustration are given in this section in order to 
enable an easier understanding of the behavior of the new measure ${\cal M}(S)$.
\\[0.5cm]
\underline{\bf First experiment.} In this experiment we illustrate the diversities 
in representing a shape as being
multi-component. We use a hand shape, displayed in Fig.\ref{saka}(a).
In the first instance, the shape in Fig.\ref{saka}(a) can be treated as a
6-component shape, whose components are the palm and five fingers, each taken as a 
separate component. In such a case the computed measure value is
${\cal M}_{6-comp.}(S)=0.0109.$

The hand shape in  Fig.\ref{saka}(a) can also be observed as a 2-component shape.
It is possible to select the palm (as displayed in Fig.\ref{saka}(b)) as one component 
and all five fingers as the other component (S displayed in Fig.\ref{saka}(c)).
In this case the computed measure is ${\cal M}_{2-comp.}(S)=0.0177.$

To complete this discussion, it is worth mentioning that if the palm 
is considered as a 1-component shape,
then ${\cal M}_{1-comp.}(S)=0,$ which is in accordance with 
the Definition \ref{definition-1},
i.e. the computed measure ${\cal M}_{1-comp.}(S)$ of all 1-component shapes is equal to zero.
If the shape consists of five fingers, as displayed in Fig.\ref{saka}(c), then the 
computed measure is ${\cal M}_{5-comp.}(S)=0.1972$.
 
\vspace*{0.3cm}

\begin{figure*}[t!]
\centering 
\hspace*{-1.5cm} 
\subfigure[$0.0626  $]{\includegraphics[width=4cm, scale=2]{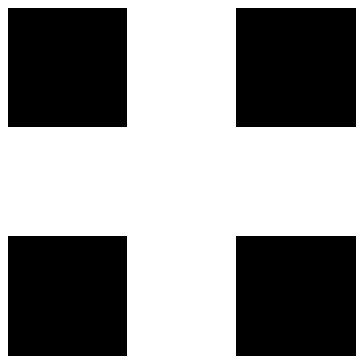}\label{Trb}}
\hspace*{-1.5cm}
\subfigure[$0.2726  $]{\includegraphics[width=4cm, scale=2]{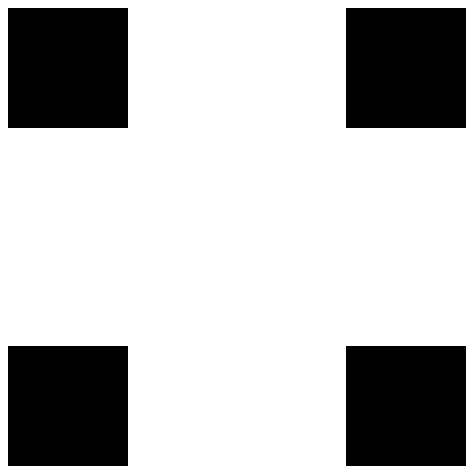}\label{Trc}}
\hspace*{-1.2cm}
\subfigure[$0.7748  $]{\includegraphics[width=4cm, scale=2]{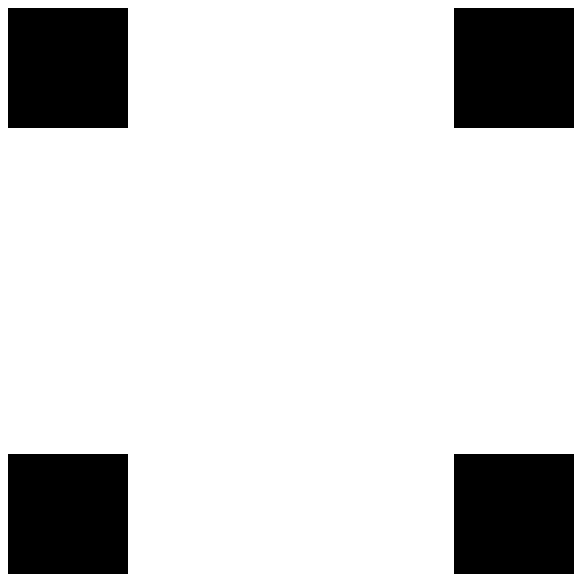}\label{Trd}}\hspace*{-0.5cm}
\subfigure[$1.8433  $]{\includegraphics[width=4cm, scale=2]{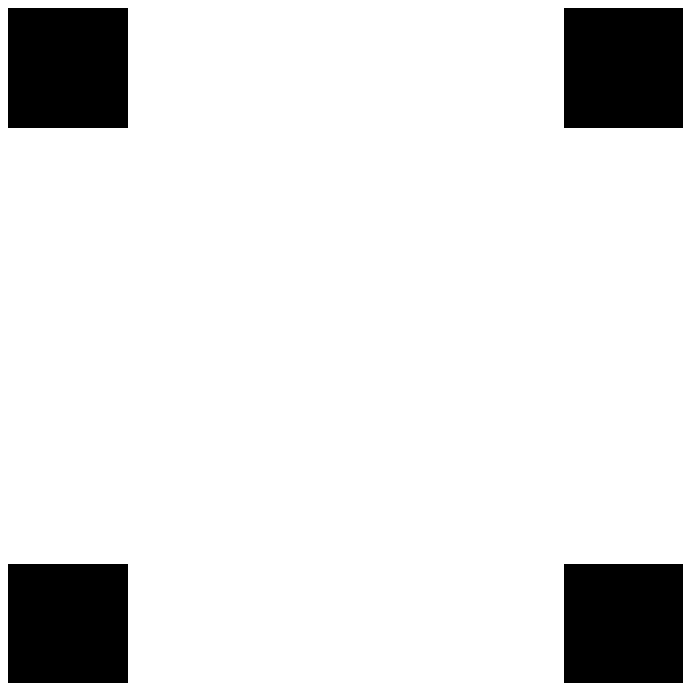}\label{Tre}}
\caption{$4$-component shapes and their assigned $ {\cal M}_{4-comp.}(S)$ values 
(given below each shape)}
\label{4-squares}
\end{figure*}

\underline{\bf Second experiment.} In this experiment we illustrate that ${\cal M}(S)$ depends 
on the distances between shape components. Precisely, as such distances increase, the 
${\cal M}(S)$ values increase as well. In each sub-figure in Fig.\ref{4-squares},
there are four identical squares, but with varying distances between each other. As the
distance between these squares increases, the measured values ${\cal M}(S)$
increase as well. Notice that this is in accordance with the definition of ${\cal M}(S)$.

\vspace*{0.3cm}

\begin{figure}[h!]
\begin{center}
\begin{tabular}{ccc}
             \includegraphics[width=3.6cm]{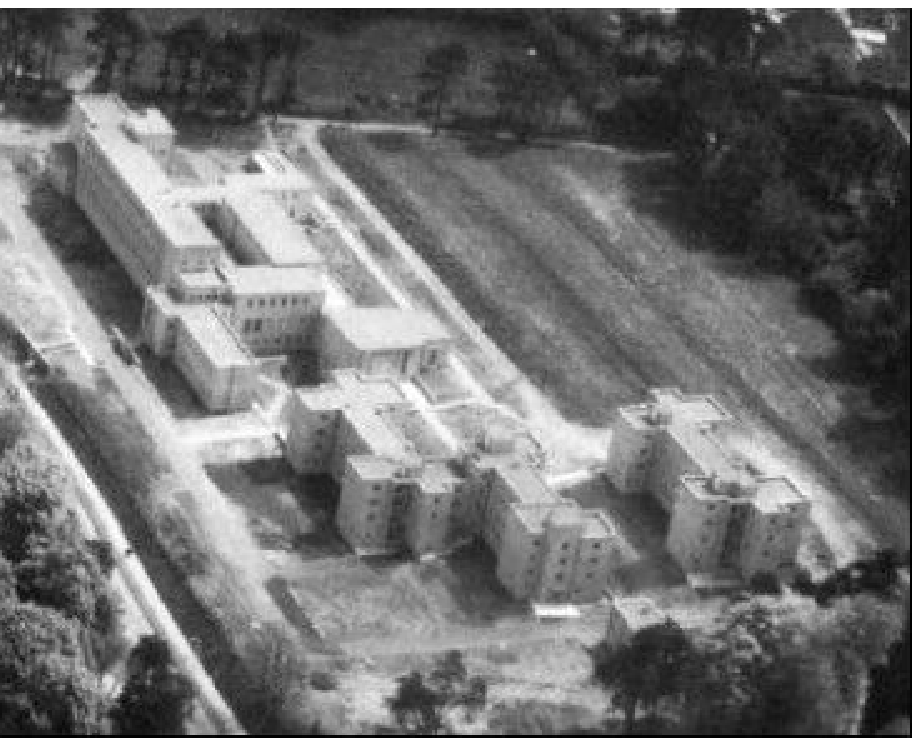}& 
					   \includegraphics[width=3.9cm]{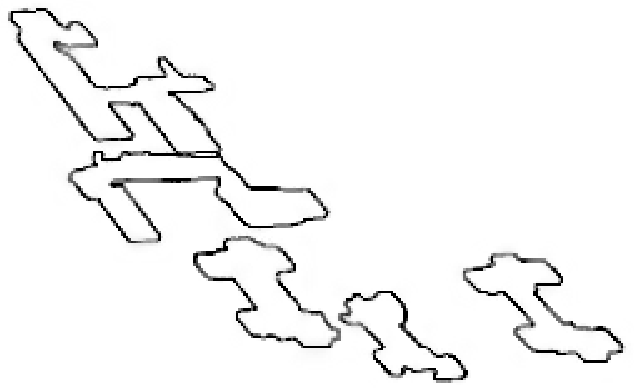}&
             \includegraphics[width=3.0cm]{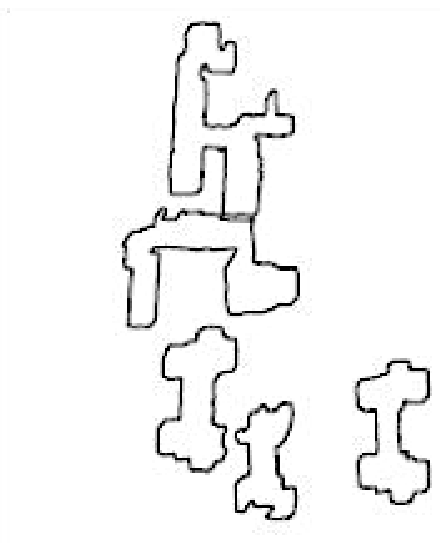}  
						\\ 
               {\footnotesize   Original image} &  
							 {\footnotesize   ${\cal M}$(S) = 0.0466} &  
							 {\footnotesize   ${\cal M}$(S) = 0.0467}    
\end{tabular}
\end{center}
\caption{The images are taken from \cite{cviu}: The original aerial image is on the left. 
The extracted contours (from the original images) are in the middle. The skewed contours,
such that the rectilinearity measure is maximized, are on the right. }
\label{cardiff}
\end{figure}

\underline{\bf Third experiment.} An aerial image (Cardiff, U.K.) is in Fig.\ref{cardiff},
on the right. The extracted  edges of the buildings are in the image in the middle. 
The obtained edges are skewed such that the rectilinearity measure 
(for detail see \cite{cviu}) of the building contours is maximized.
The obtained values for the two 5-component shapes are almost identical 
(${\cal M}(middle-image)=0.0324$ and ${\cal M}(image-on-the-right)$), as expected.
There is a small difference caused by the numerical calculation.   
 
\vspace*{0.3cm}

\begin{figure}[h!]
\begin{center}
\begin{tabular}{ccc}
             \includegraphics[width=3.0cm]{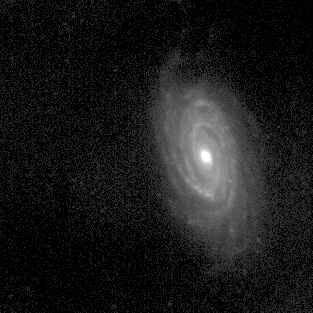}& 
					   \includegraphics[width=3.0cm]{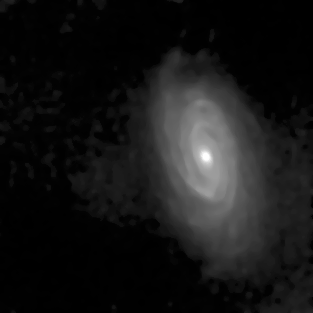}& 
						 \includegraphics[width=3.0cm]{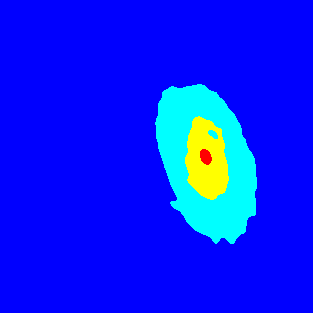} \\ 
						  & & $ {\cal M}_{3-comp.}(S) = 0.000679  $   \\  
               {\footnotesize   Original image} &  
							 {\footnotesize   Filtered image} &  
							 {\footnotesize   Multicomponent image}     \\ \hline \\
						 \includegraphics[width=3.0cm]{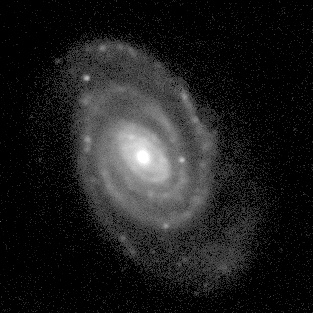}& 
					   \includegraphics[width=3.0cm]{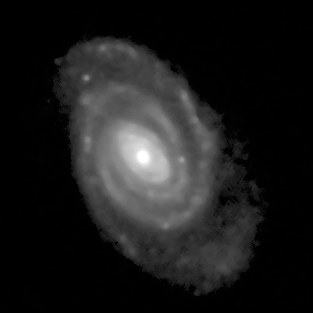}& 
						 \includegraphics[width=3.0cm]{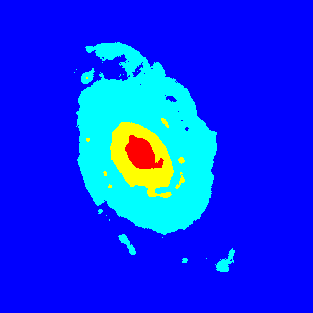} \\
						& & $  {\cal M}_{3-comp.}(S)  = 0.000593  $    
						\\  
               {\footnotesize   Original image} &  
							 {\footnotesize   Thresholded image} &  
							 {\footnotesize   Multicomponent image}    
							 \\   \hline   \\ 
						 \includegraphics[width=3.0cm]{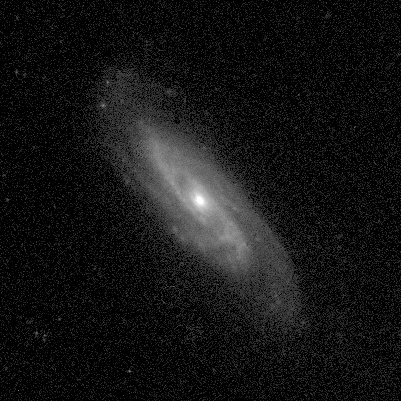}& 
					   \includegraphics[width=3.0cm]{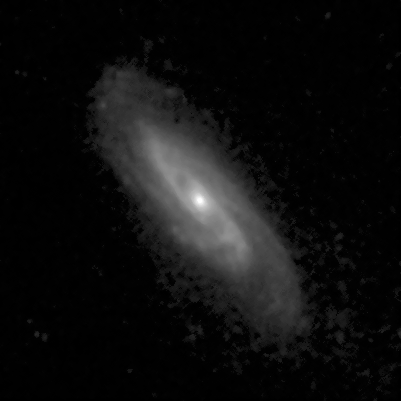}& 
						 \includegraphics[width=3.0cm]{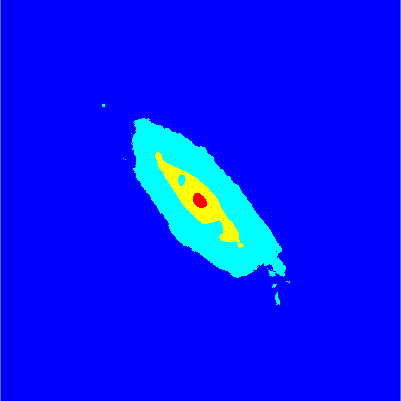} \\
						 & &  ${\cal M}_{3-comp.}(S) =  0.001883  $    
						\\  
               {\footnotesize   Original image} &  
							 {\footnotesize   Filtered image} &  
							 {\footnotesize   Multi-component image}  \\ \hline  
\end{tabular}
\end{center}
\caption{The original images of the three spiral galaxies, from \cite{frei}, are displayed 
on the left, in each row. The images in the second column, are obtained by 
applying a median filter, to the original galaxy images. The images on the right are 
4-component images, obtained by a thresholding method. Three components are used for the
${\cal M}_{3-comp}(S)$ computation, while the fourth component corresponds to the 
galaxy background.}
\label{spiral}
\end{figure}

\begin{figure}[h!]
\begin{center}
\begin{tabular}{ccc}
             \includegraphics[width=3.0cm]{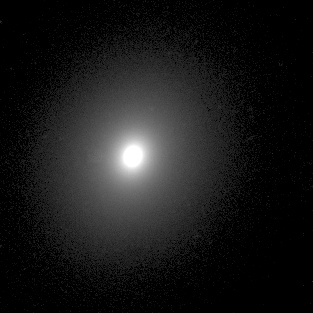}& 
					   \includegraphics[width=3.0cm]{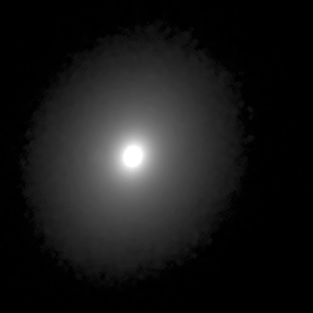}& 
						 \includegraphics[width=3.0cm]{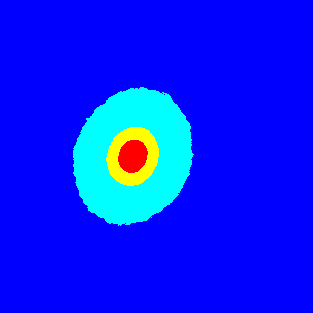} \\
						& &  ${\cal M}_{3-comp.}(S) = 0.000443 $    
						\\  
               {\footnotesize   Original image} &  
							 {\footnotesize   Filtered image} &  
							 {\footnotesize   Multicomponent image} \\   
							  \hline   \\
						 \includegraphics[width=3.0cm]{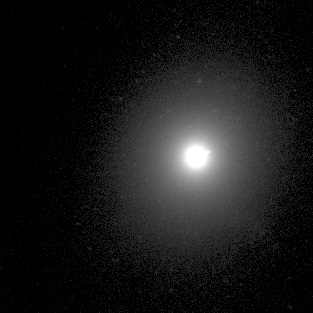}& 
					   \includegraphics[width=3.0cm]{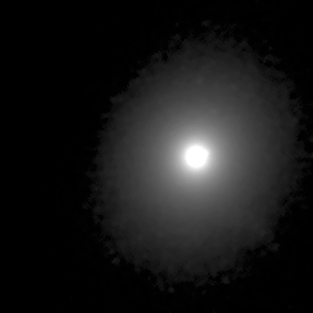}& 
						 \includegraphics[width=3.0cm]{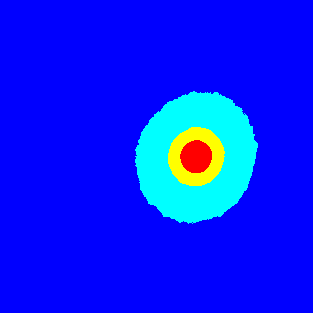} \\ 
						 & & $ {\cal M}_{3-comp.}(S) =  0.000545 $    
						\\ 
               {\footnotesize   Original image} &  
							 {\footnotesize   Thresholded image} &  
							 {\footnotesize   Multicomponent image} \\ \hline \\
						 \includegraphics[width=3.0cm]{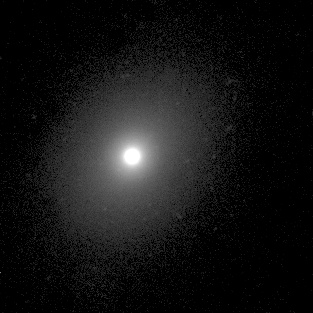}& 
					   \includegraphics[width=3.0cm]{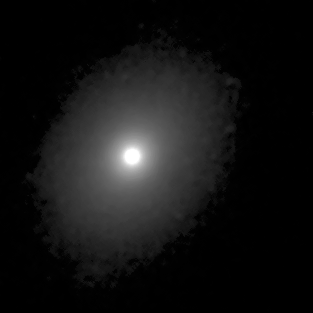}& 
						 \includegraphics[width=3.0cm]{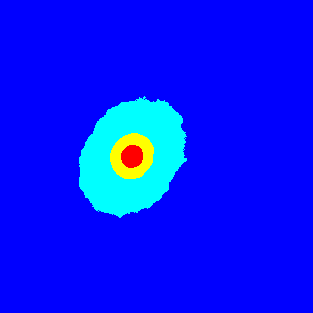} \\
						 & & ${\cal M}_{3-comp.}(S) = 0.000341  $    
						\\ 
               {\footnotesize   Original image} &  
							 {\footnotesize   Filtered image} &  
							 {\footnotesize   Multicomponent image}  \\ \hline  
\end{tabular}
\end{center}
\caption{The original images of the three elliptical galaxies, from \cite{frei}, are displayed 
on the left, in each row. The images in the second column, are obtained by 
applying a median filter, to the original images. The images on the right are 
4-component shapes, obtained by a thresholding method. Three components are used for the
${\cal M}_{3-comp}(S)$ computation, while the fourth component corresponds to the 
galaxy background.}
\label{elliptical}
\end{figure}

\underline{\bf Fourth experiment.}
In this experiment we illustrate the potential of the new measure to be 
used in a galaxy shape analysis task. Elliptical and spiral galaxies, 
listed in the popular Nearby Galaxy Catalog \cite{frei}, are selected to illustrate 
a discriminative capacity of ${\cal M}_{3-comp.}(S)$. The results for spiral galaxies 
are in Fig.\ref{spiral}, while the results for elliptical galaxies are in 
Fig.\ref{elliptical}.
The original galaxy images are on the left, in each row, in both Fig.\ref{spiral} and
Fig.\ref{elliptical}. A median filter has been applied to these original images. 
The filtered images are in the second column, in both related images. 
Multi-components of the filtered images are obtained by applying a multi threshold method.
These images are on the right.  
Four components were obtained for each galaxy filtered image. 
Three of them are used as galaxy image components, 
and the fourth component corresponds to the galaxy background. 
The computed ${\cal M}_{3-comp.}(S)$ 
measure is immediately below the thresholded images.
The measure  ${\cal M}_{3-comp.}(S)$ distinguishes between spiral and elliptical 
galaxies very well. The values for ${\cal M}_{3-comp.}(S)$ are $0.000679$ 
(for the galaxy image in 
the first row), $0.000593$ for the galaxy image in the second row, and $0.001883$ 
for the galaxy image
in the third row (see Fig.\ref{spiral}). The computed ${\cal M}_{3-comp.}(S)$ values are smaller for
the three selected elliptical galaxies. They are $0.000545,$ $0.000341,$ 
and $0.000341$, respectively. 

\section{Conclusion} 
A new shape measure ${\cal M}(S)$ has been introduced. 
The new measure is Affine invariant and is designed to 
be applicable to multi-component shapes. The measure is developed by exploiting a new 
geometric interpretation of the well-known Affine invariant \cite{suk}:
$\ {\cal A}(S)   =   \mu_{2, 0}(S) \cdot  \mu_{0, 2}(S) - \mu_{1, 1}(S)^2 \ $ (see (\ref{A-def})). 
This new interpretation of ${\cal A}(S)$  says that ${\cal A}(S)$ is proportional 
to the total integral of the squared areas of all the triangles whose vertices belong to $S$.
Another interpretation of ${\cal A}(S)$ can be found in \cite{xu}, but this 
interpretation does not allow an easy application to the multi-components shapes measurement. 
The new interpretation of
${\cal A}(S)$ enables an easy and natural extension of ${\cal A}(S)$ to multi-component shapes.
As such, the new measure ${\cal M}(S)$ is very flexible, because the concept of multi-component 
shapes is very generic. Indeed, a given shape can be decomposed into components in various ways.
Also, the number of components can be selected in different ways. All this has been illustrated on 
illustrative experimental examples.  

The new measure ${\cal M}(S)$ is an Affine invariant. The suitability of the Affine invariance property
for multi-component shape analysis is also illustrated. The task of matching aerial photos of objects on the
ground (please see the third experiment) was given as an illustration.
 
Being an area based measure, i.e. a measure that uses all of the shape points for its computation, 
it is efficient to compute \cite{huxley,klette}. This is not a case for 
boundary based shape measures, e.g. \cite{anisotropy,milos,anastazia}.

\section*{References}

\bibliography{ref}{}

\end{document}